\documentclass[10pt]{article}

\usepackage[margin=0.8in]{geometry}
\usepackage{times}
\usepackage{microtype}
\usepackage{graphicx}
\usepackage{booktabs}
\usepackage{multirow}
\usepackage{amsmath}
\usepackage{url}
\usepackage[hidelinks]{hyperref}
\usepackage[numbers,sort&compress]{natbib}
\usepackage{xcolor}
\usepackage{caption}
\usepackage{subcaption}
\title{Deep Learning-Based Prediction of HER2 Status from Whole-Brain MRI in Breast Cancer Brain Metastases}

\author{
Sara Early$^{1,2,3,4}$,
Chris Kang$^{1,2,3}$,
Vibujithan Vigneshwaran$^{1,2,3}$,\\
Kimberly Amador$^{1,2,3}$,
Roger Y. Tsang$^{5, 6,7}$,
Nils D. Forkert$^{1,2,3,5,8}$\\[4pt]
\parbox{0.95\textwidth}{
\centering
\footnotesize
$^1$Department of Radiology, University of Calgary, Calgary, AB, Canada\\
$^2$Hotchkiss Brain Institute, University of Calgary, Calgary, AB, Canada\\
$^3$Alberta Children's Hospital Research Institute, University of Calgary, Calgary, AB, Canada\\
$^4$Department of Biomedical Engineering, University of Calgary, Calgary, AB, Canada\\
$^5$Annie Charbonneau Cancer Institute, University of Calgary, Calgary, AB, Canada\\
$^6$Department of Oncology, Cumming School of Medicine, University of Calgary, Calgary, AB, Canada\\
$^7$Arthur J.E. Child Comprehensive Cancer Centre, Calgary, AB, Canada\\
$^8$Department of Clinical Neurosciences, Cumming School of Medicine, University of Calgary, Calgary, AB, Canada
}}

\date{}

\begin{document}
\maketitle

\begin{abstract}

\textbf{Purpose} Human epidermal growth factor receptor 2 (HER2) status is a key biomarker relevant for prognostication and treatment decision making in metastatic breast cancer. Thus, the aim of this work was to explore whether HER2 status can be predicted directly from whole-brain post-contrast T1-weighted (T1w) MRI scans in breast cancer brain metastases (BCBM) using deep learning, and whether providing explicit tumour segmentation information improves prediction. 

\textbf{Methods} Using the publicly available BCBM-Radiogenomics dataset, we analysed 135 patients, including 70 HER2+ and 65 HER2- patients. Three approaches were implemented and evaluated: a conventional radiomics model, a simple fully convolutional network (SFCN) using T1w MRI scans alone, and an SFCN using T1w MRI scans with an expert-derived tumour mask as an additional input channel. 

\textbf{Results} The radiomics model achieved an area under the receiver operating characteristic curve (ROC-AUC) of 0.602 (95\% confidence interval [CI]: 0.528--0.677) while the SFCN achieved a ROC-AUC of 0.667 (95\% CI, 0.63--0.71) using T1w MRI scans alone, which increased to 0.692 (95\% CI, 0.58--0.81) when an explicit tumour mask was additionally provided. 

\textbf{Conclusion} However, the modest performance across all approaches demonstrates the difficulty of predicting HER2 status from structural MRI scans alone. The difference between the segmentation-free and mask-guided SFCN models suggests that much of the predictive signal captured by the deep-learning model may be learned directly from whole-brain T1w MRI without requiring explicit tumour delineation. These findings support further investigation of segmentation-free deep-learning approaches as a potential non-invasive strategy to support HER2 status determination in BCBM.
\end{abstract}


\section{Introduction}

Human epidermal growth factor receptor 2 (HER2) is a key biomarker in breast cancer that provides prognostication and directly influences treatment selection, particularly in metastatic disease \citep{zimmer2022her2}. Brain metastases are a clinically significant and major complication of advanced breast cancer, occurring in up to 50\% of HER2+ cases \citep{ramakrishna2022management}. In those cases, the HER2 status is particularly relevant because it can influence eligibility for HER2-directed systemic therapies. Although HER2 status is conventionally determined from tumour tissue sampling at the primary site, the receptor phenotype is not necessarily conserved during metastatic progression. A systematic review and meta-analysis of 15 studies including 1,373 patients reported a HER2 discordance between primary breast tumours and breast cancer brain metastases (BCBM) in approximately 12\% of patients, including both HER2 gain and loss \citep{kotecha2021discordance}. Furthermore, a recent study has shown supporting evidence for discordance in HER2 expression between primary breast tumours and brain metastases, including changes between HER2+ and HER2-, as well as between HER2-low and HER2-zero subcategories of HER2- \citep{pereslete2025analysis}. Consequently, the HER2 status determined from the primary tumour may not always accurately represent the molecular phenotype of contemporary intracranial disease.

Tissue-based assessment remains the gold standard for determining the receptor status, and clinical guidelines recommend biopsy and retesting of estrogen receptor, progesterone receptor, and HER2 status when accessible metastatic tissue is available \citep{van2015biomarkers}. However, obtaining tissue from a brain metastasis requires an invasive procedure and may not be clinically indicated or feasible in every patient. In the absence of contemporary metastatic tissue, clinicians may decide to rely on previously obtained receptor information, which may not be accurate for the BCBMs. This creates a potential role for non-invasive imaging approaches that could provide complementary information regarding the molecular characteristics of BCBMs.

Radiogenomics provides a framework for investigating whether quantitative imaging characteristics are associated with tumour biology. Prior studies associating receptor status in BCBM with imaging features have predominantly used lesion-based radiomic approaches, extracting predefined quantitative features from segmented metastatic lesions \citep{luo2022radiomics,heitkamp2023receptor}. Overall, these studies suggest that MRI scans contain potentially useful information related to receptor status, but conventional radiomic approaches depend on explicit tumour delineation and traditional image feature filters that may not work well in case of small lesion and may miss important image features. This may be especially problematic in patients with BCBM, who frequently present with multiple brain metastases of different sizes. 

More recently, a few deep learning approaches have been investigated for receptor prediction from BCBM MRI scans. Therefore, deep-learning-derived features have been extracted from defined intratumoural regions \citep{shi2022deep}, while a multimodal framework combining 3D convolutional features, radiomics, and anatomical position used segmentation-derived lesion-centred image regions for simultaneous prediction of estrogen receptor, progesterone receptor, and HER2 status \citep{sun2026deep}. Segmentation-free prediction has also recently been explored using a 3D convolutional network trained directly on post-contrast T1-weighted (T1w) MRI brain scans \citep{christodoulou2026saliency}. Collectively, these studies demonstrate increasing interest in utilizing learned MRI representations for BCBM receptor prediction, but differ substantially in their dependence on explicit lesion segmentation. This variation raises an important question: Does providing tumour-segmentation information offer a meaningful predictive advantage in whole-brain deep-learning models and if and to what extent do they outperform conventional radiomics methods. In a segmentation-free configuration, a whole-brain deep learning model must learn both where potentially relevant information is located and which imaging characteristics are associated with HER2 status. In contrast, a tumour mask helps the model to focus on metastatic disease areas while also providing direct information regarding lesion extent, morphology, burden, and spatial distribution. However, while advantageous in theory, requiring such a mask at the inference stage introduces an additional workflow dependency. Thus, a direct comparison of otherwise matched models with and without tumour-mask information can help to determine whether and to what extent metastases segmentations provide additional predictive value over whole-brain T1w MRI scans alone.

In this proof-of-concept study, we investigated whether patient-level HER2 status can be predicted from post-contrast T1w MRI scans in patients with clinically validated BCBM. We evaluated three complementary approaches: (1) a conventional radiomics model, (2) a simple fully convolutional network (SFCN) using whole-brain T1w MRI scans alone, and (3) an otherwise matched SFCN using T1w MRI scans with an expert-delineated tumour mask as an additional input channel. The primary objectives were to determine whether a whole-brain deep-learning model can capture discriminative information associated with HER2 status and whether and to what extent explicit tumour-segmentation information improves the prediction relative to a segmentation-free deep learning model.

\section{Materials and Methods}

\begin{figure}[htbp]
    \centering
    \includegraphics[width=0.7\textwidth]{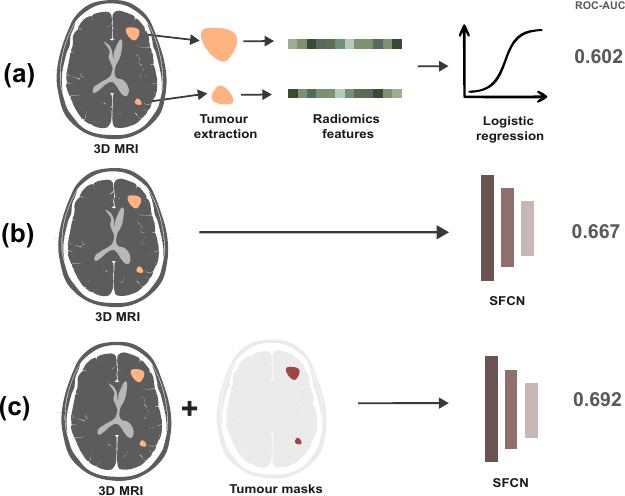}
    \caption{Three approaches were evaluated for HER2 status classification: (a) a conventional radiomics model, (b) a simple fully convolutional network (SFCN) using T1w MRI scans alone, and (c) an SFCN using T1w MRI scans with an expert-delineated tumour mask as an additional input channel}
    \label{fig:ga}
\end{figure}

An overview of the three main approaches described in this work is shown in Fig.~\ref{fig:ga}.

\subsection{Dataset, study population, and image preprocessing}

Data were obtained from the publicly available BCBM-Radiogenomics dataset \citep{taha2025bcbm}, which contains post-contrast T1w brain MRI scans from patients with breast cancer and metastases in the brain, together with expert-delineated lesion segmentations and patient-level receptor status information (see below). The segmented images included in this dataset were produced via a semi-automated, consensus segmentation method. A board-certified neurosurgeon and a radiation oncologist utilized segmentation software within the stereotactic radiosurgery system for slice-by-slice delineation of the metastatic brain tumors and neighboring structures at risk.

According to the dataset description, histologic information was obtained from lymph-node biopsy, primary-breast sampling, or biopsy of a metastatic site, which could be intracranial or extracranial. Therefore, the publicly released HER2 label represents a patient-level clinical receptor-status label derived from available tissue sampling and does not necessarily reflect the HER2 status of the imaged brain metastases. Patient-specific information regarding the tissue source used for HER2 determination, the timing of tissue sampling relative to MRI, and the assay criteria used to assign HER2 status were not available in the released dataset. We will therefore apply standard ASCO/CAP criteria to define HER2+ status, which is most commonly characterized by IHC 3+ or IHC 2+ with positive ISH testing.

For the present patient-level analysis, only patients with a reported binary HER2 status were included, and the earliest available post-contrast T1w MRI scan was retained for each patient. The final cohort included 135 patients, comprising 70 HER2+ and 65 HER2- patients. The demographic, disease-burden, and MRI acquisition characteristics of the study cohort are shown in Table~\ref{tab:cohort}.

\begin{table}[ht]
\centering
\caption{MRI acquisition characteristics of the study cohort stratified by HER2 status.}
\label{tab:cohort}
\renewcommand{\arraystretch}{1.0}
\begin{tabular}{lccc}
\toprule
Cohort variable & Overall ($n=135$) & HER2+ ($n=70$) & HER2- ($n=65$) \\
\midrule
Age, years (mean $\pm$ SD)
& 56.9 $\pm$ 10.4
& 56.6 $\pm$ 10.4
& 57.2 $\pm$ 10.5 \\
Number of metastases (mean $\pm$ SD; median)
& 3.3 $\pm$ 2.9; 2
& 3.1 $\pm$ 2.9; 2
& 3.4 $\pm$ 2.8; 2 \\
Total lesion volume, cm$^3$ (mean $\pm$ SD; median)
& 6.7 $\pm$ 8.3; 3.6
& 5.8 $\pm$ 8.9; 3.1
& 7.6 $\pm$ 7.5; 5.3 \\
1.5T scans, $n$
& 126 
& 63 
& 63 \\
3T scans, $n$
& 9
& 7
& 2 \\
\bottomrule
\end{tabular}
\end{table}

Scans were acquired on Siemens systems at 1.5T (TR/TE, 9.4/4.76~ms; flip angle, $25^\circ$) or 3T (TR, 1900~ms; TE, 2.49--3.46~ms; flip angle, $12^\circ$), with native voxel spacing of $0.5\times0.5\times1.0$~mm$^3$ \citep{taha2025bcbm}. Individual scanner models and identifiers were not reported.

All post-contrast T1w MRI scans underwent skull-stripping using HD-BET \citep{isensee2019hdbet}, followed by N4 bias-field correction \citep{tustison2010n4}, z-score intensity normalization, non-zero image cropping, and resampling to $1\times1\times1$~mm$^3$ isotropic resolution. Volumes were resized to a fixed network input size of $160\times160\times144$ voxels using trilinear interpolation. 

\subsection{Radiomics}
\label{subsec:radiomics}

A conventional radiomics model was developed and evaluated for HER2 prediction as a baseline for HER2 prediction. The dataset provides 107 radiomic features extracted from expert-delineated segmentations using PyRadiomics, comprising shape, first-order intensity, and texture features \citep{taha2025bcbm}. Because patients frequently present with multiple metastases, lesion-level radiomic features were aggregated to the patient level using the largest lesion's features or mean, max, or combined mean and max pooling across lesions. Tumour-burden measures and patient age were also evaluated (Table~\ref{tab:radiomics_representations}).

\begin{table}[ht!]
\centering
\caption{Patient-level feature representations evaluated for HER2 classification.}
\label{tab:radiomics_representations}
\renewcommand{\arraystretch}{1.1}
\begin{tabular}{lcp{8.0cm}}
\toprule
Feature representation & No. of features & Definition \\
\midrule

Tumour burden
& 4
& Number of lesions, total lesion volume, largest lesion volume, and log-transformed total lesion volume. \\

Largest-lesion radiomics
& 107
& Complete radiomic feature vector from the largest lesion for each patient. \\

Mean-pooled radiomics
& 107
& Mean of each radiomic feature across all lesions for each patient. \\

Max-pooled radiomics
& 107
& Maximum value of each radiomic feature across all lesions for each patient. \\

Mean- and max-pooled radiomics
& 214
& Concatenation of the mean- and max-pooled radiomic feature vectors. \\

Largest-lesion radiomics + age
& 108
& Largest-lesion radiomic feature vector combined with patient age. \\

Age only
& 1
& Patient age used as a non-imaging baseline. \\

\bottomrule
\end{tabular}
\end{table}

For each of the seven representations, a logistic regression classifier with feature standardization was trained using an L1 or L2 penalty, with the penalty type and regularization strength $C \in \{10^{-3},10^{-2},10^{-1},1,10\}$ selected via nested cross-validation: within each of the five outer folds, hyperparameters were chosen using an inner five-fold split of the training data only, and the selected configuration was then applied to generate an out-of-fold prediction for the corresponding held-out outer fold.

The primary performance metric was ROC-AUC, reported with a bootstrap 95\% confidence interval. The tumour-burden representation, which achieved the highest nested ROC-AUC and was the only representation for which the permutation test reached $p<0.05$, was selected as the radiomics baseline reported alongside the SFCN models. The full seven-way comparison is reported in Table~\ref{tab:radiomics_features}.

\subsection{Deep-learning input configurations}

Two input configurations were evaluated using the SFCN architecture:
\begin{itemize}
    \item \textbf{T1w MRI scans without tumour-segmentation information}. The preprocessed whole-brain post-contrast T1w MRI scan was provided directly to the model as a single-channel input without lesion masking. This configuration does not require any tumour segmentation at inference and represents the lowest-workflow input configuration evaluated.
    \item \textbf{T1w MRI with tumour-segmentation information}. Expert-delineated metastatic lesion segmentations were combined into a single patient-level tumour mask and supplied to the model as a second input channel alongside the full, unmodified T1w MRI scan. Tumour voxels were not removed, zeroed, or replaced. This comparison was designed to determine whether providing explicit tumour segmentation information improves prediction beyond what can be achieved using whole-brain T1w MRI scans alone.
\end{itemize}

\subsection{Simple Fully Convolutional Network}

The simple fully convolutional network (SFCN) \citep{peng2021sfcn} consists of six convolutional blocks with channel widths of 32, 64, 128, 256, 256, and 64, followed by adaptive global average pooling, dropout with probability 0.5, and a single linear output for binary classification.

The T1w-only SFCN model used a single input channel, whereas the mask-guided model used two input channels corresponding to the T1w MRI scan and the corresponding tumour mask. Both models were trained identically in all other respects, using the binary cross-entropy loss, the Adam optimizer with an initial learning rate of $1\times10^{-4}$, batch size 2, and a maximum of 100 epochs, with early stopping based on the validation AUROC (patience of 20 epochs). Class imbalance was addressed using weighted random sampling. No learning-rate scheduling, gradient clipping, or positive-class loss re-weighting was applied to either model. Thus, the two configurations differed only in the number of input channels. All trainings utilized fixed seeds.

\subsection{Experimental design and cross-validation}

For the evaluation of the three different models implemented and tested, the 135 patients were divided into five folds by stratified sampling on HER2 status at the patient level (giving 27 patients per fold, 14 HER2-positive and 13 HER2-negative in each). For every fold in turn, that fold was held out as the test set and the remaining 108 patients were further divided 80/20, again stratified on HER2, into 86 training and 22 validation patients. All 435 lesions inherit their patient's fold and train/validation/test assignment rather than being assigned independently, so that a patient's lesions can never be divided across a split. Model hyperparameters, the stopping epoch for the networks, and the choice of per-patient aggregation were selected on validation patients only, and each held-out fold was scored exactly once (Section~\ref{subsec:evaluation}).

The classification threshold was fixed at 0.5 without post-hoc threshold optimization. Within each fold, the held-out fold was used for model selection and final performance evaluation.

\subsection{Evaluation metrics and statistical tests}
\label{subsec:evaluation}

Area under the receiver operating characteristic curve (ROC-AUC) was used as the primary performance metric, where the HER2+ status was treated as the positive class. Radiomics and SFCN approaches were assessed using two-sided paired $t$-tests on the five fold-level ROC-AUC values ($df=4$): T1w SFCN \textit{vs.} T1w+mask SFCN, T1w SFCN \textit{vs.} radiomics, and T1w+mask SFCN \textit{vs.} radiomics. A significance threshold of $\alpha=0.05$ was used.

Experiments ran on a single NVIDIA RTX 3090 (24 GB) with an Intel i7-11700F and 61 GB RAM, using Python 3.9, PyTorch 2.8 (CUDA 12.8) and scikit-learn 1.6. The radiomics experiments are CPU-only and take seconds, while the deep learning models need the GPU, roughly 120s per fold for the whole-brain SFCN.

\section{Results}

\subsection{Comparison of radiomics feature representations}

Table~\ref{tab:radiomics_features} summarizes nested ROC-AUC for all seven radiomics feature representations described in Section~\ref{subsec:radiomics}. The tumour-burden representation (4 features) achieved the highest ROC-AUC (0.602). Representations incorporating full shape and texture radiomics (107--214 features) performed no better than tumour burden despite substantially higher dimensionality, while the clinical-only (age) baseline performed at chance level.

As noted in Section~\ref{subsec:radiomics}, tumour burden was selected as the reported radiomics baseline because it was the best-performing of the seven representations evaluated. Since its permutation $p$-value of 0.034 does not survive correction for this seven-way comparison (Bonferroni $\alpha=0.05/7\approx0.007$), it should be interpreted as exploratory rather than confirmatory.

\begin{table}[ht]
\centering
\caption{Patient-level HER2 classification performance of the seven feature representations evaluated using nested cross-validation. ROC-AUC, area under the receiver operating characteristic curve; CI, confidence interval.}
\label{tab:radiomics_features}
\renewcommand{\arraystretch}{1.0}
\setlength{\tabcolsep}{6pt}
\begin{tabular}{lccc}
\toprule
Feature representation & No. of features & ROC-AUC & 95\% CI \\
\midrule
Tumour burden            & 4   & \textbf{0.602} & 0.528--0.677 \\
Largest-lesion radiomics & 107 & 0.553 & 0.460--0.646 \\
Mean+max radiomics       & 214 & 0.552 & 0.451--0.653 \\
Mean-pooled radiomics    & 107 & 0.541 & 0.422--0.659 \\
Max-pooled radiomics     & 107 & 0.536 & 0.431--0.642 \\
Largest-lesion + clinical& 108 & 0.525 & 0.305--0.745 \\
Clinical only (age)      & 1   & 0.461 & 0.390--0.532 \\
\bottomrule
\end{tabular}
\end{table}

\subsection{Comparison between radiomics and deep-learning approaches}

\begin{figure}[htbp]
    \centering
    \includegraphics[width=0.65\textwidth]{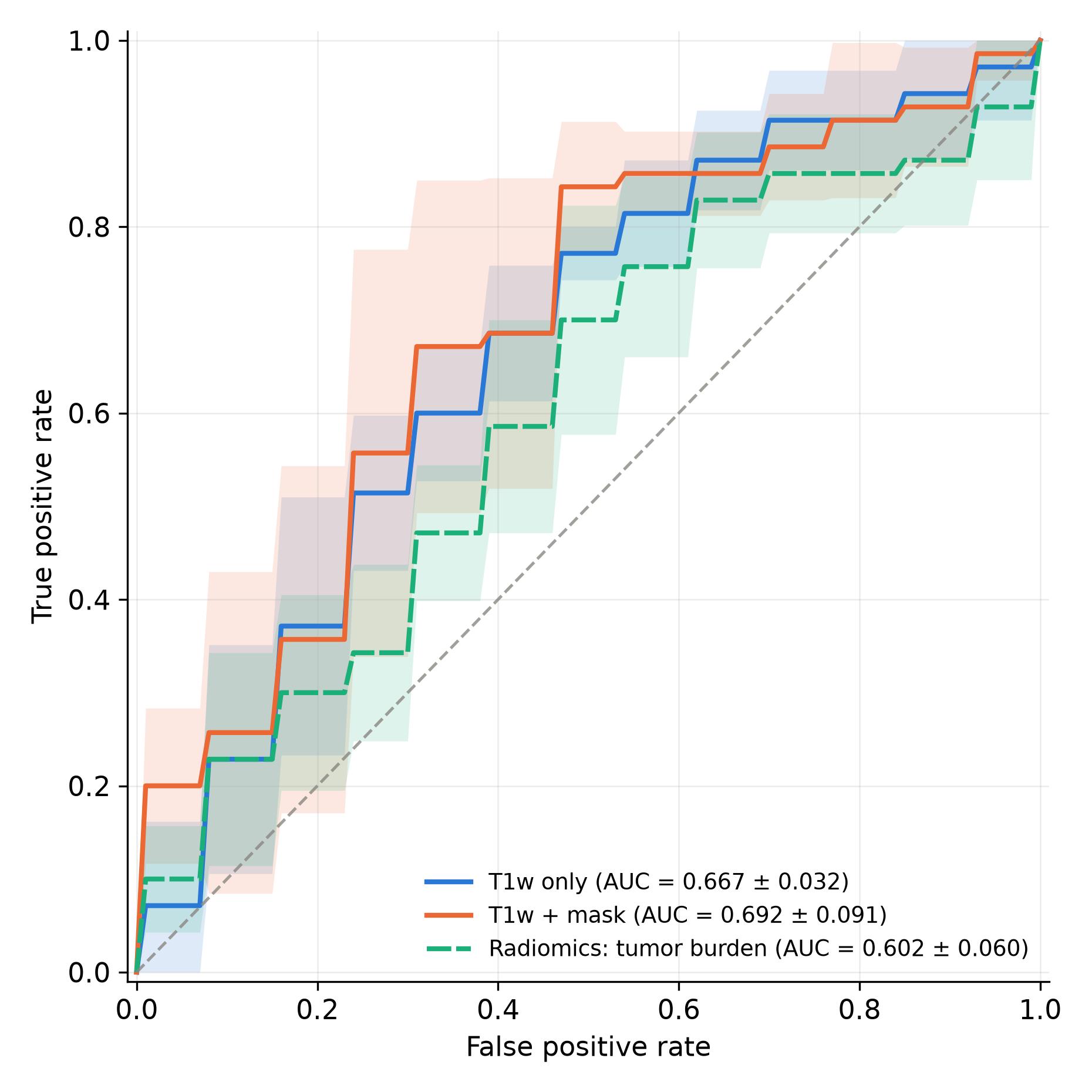}
    \caption{Receiver operating characteristic (ROC) curves for HER2 status prediction. Curves are shown for the tumour-burden radiomics model, the T1w-only SFCN, and the T1w + tumour mask SFCN. The corresponding mean-of-folds AUCs were 0.602, 0.667, and 0.692, respectively}
    \label{fig:her2_roc_curves}
\end{figure}

Table~\ref{tab:performance} summarizes HER2 classification performance across the conventional radiomics baseline and the two SFCN input configurations; ROC curves for all three approaches are shown in Fig.~\ref{fig:her2_roc_curves}. The tumour-burden radiomics model achieved a ROC-AUC of 0.602 (95\% CI: 0.528--0.677). Both SFCN models outperformed this baseline on point estimate, achieving 0.667 (95\% CI, 0.63--0.71) using T1w MRI scans alone, slightly improving to 0.692 (95\% CI, 0.58--0.81) with the tumour mask as an additional input. Neither difference reached statistical significance (paired $t$-tests on fold-level AUC, Section~\ref{subsec:evaluation}: T1w-only \textit{vs.}\ radiomics, $t(4)=1.94$, $p=0.13$; T1w+mask \textit{vs.}\ radiomics, $t(4)=1.46$, $p=0.22$), which could be read as a directional advantage for the deep-learning approach rather than confirmed superiority.

\begin{table}[ht]
\centering
\caption{Comparison of radiomics and SFCN approaches for patient-level HER2 classification performance.}
\label{tab:performance}
\renewcommand{\arraystretch}{1.0}
\setlength{\tabcolsep}{6pt}
\begin{tabular}{lllcc}
\toprule
Model & Input & Segmentation required & ROC-AUC & 95\% CI \\
\midrule
Radiomics (Logistic Regression) & 4 radiomic features
& Yes
& 0.602
& 0.528--0.677 \\

SFCN
& T1w MRI & No
& 0.667
& 0.63--0.71 \\

SFCN
& T1w MRI + tumour mask & Yes
& 0.692
& 0.58--0.81 \\
\bottomrule
\end{tabular}
\end{table} 

The addition of an explicit tumour mask produced only a modest, non-significant increase in SFCN performance, from a ROC-AUC of 0.667 using T1w MRI scans alone to 0.692 with mask guidance (Table~\ref{tab:performance}; absolute increase of 0.025; paired $t$-test on fold-level AUC, $t(4)=0.64$, $p=0.56$). This finding suggests that providing explicit tumour segmentation information may offer additional predictive value. However, the model retained measurable discriminatory performance when trained directly on the whole-brain T1w MRI scans without a tumour mask. 

\section{Discussion}

In this work, we investigated whether HER2 status can be predicted from post-contrast T1w MRI scans in patients included in the BCBM-Radiogenomics dataset. Therefore, we compared a conventional tumour-burden radiomics model with two SFCN models that used either whole-brain T1w MRI scans alone or T1w MRI scans with an explicit tumour mask. All three approaches demonstrated modest discrimination, with ROC-AUCs of 0.602, 0.667, and 0.692, respectively. Although none of the pairwise differences reached statistical significance, the SFCN using whole-brain T1w MRI alone achieved a performance similar to the mask-guided SFCN, with only a 0.025 difference in AUC. This finding provides preliminary evidence that a deep-learning model may capture HER2-associated imaging information from whole-brain MRI without the need for expert segmentation input.

The clinical motivation for exploring this problem arises from the difficulty of acquiring contemporary molecular information from brain metastases. HER2 status can differ between the primary breast tumour and later stage BCBM, with a meta-analysis reporting approximately 12\% discordance of HER2 status between primary tumours and brain metastases \citep{kotecha2021discordance}. Current clinical practice relies heavily on tissue-based assessment of receptor status. Clinical guidelines recommend biopsy and retesting of estrogen receptor, progesterone receptor, and HER2 status when accessible metastatic tissue is available \citep{van2015biomarkers,wolff2018her2}. However, intracranial biopsy or resection is invasive and is not performed solely for molecular characterization in every patient. Therefore, an imaging-based approach would be advantageous as a complementary source of information when a tissue biopsy is impractical or unavailable.

Prior studies investigating non-invasive prediction of the receptor status in BCBM have predominantly used radiomic features extracted from delineated metastatic lesions \citep{luo2022radiomics,heitkamp2023receptor}. These approaches have demonstrated that MRI scans contain information associated with receptor phenotype, but require a segmented tumour region before radiomic features can be extracted. In contrast, the T1w-only SFCN presented in this work was provided with the complete preprocessed brain volume and no tumour mask, allowing HER2 prediction without explicit tumour segmentation. The T1w-only SFCN achieved a ROC-AUC of 0.667, compared with 0.602 for the tumour-burden radiomics baseline model. Although this difference was not statistically significant, the result provides preliminary evidence that a learned whole-brain representation may capture relevant information beyond simple measures of tumour burden.

The comparison between the two SFCN input configurations provides an important finding. Adding the tumour mask increased the ROC-AUC slightly from 0.667 to 0.692, an absolute improvement of 0.025. Thus, although the addition of the tumour mask moderately improved the classification performance, these results show that the segmentation-free T1w model was able to achieve a similar discrimination for predicting HER2 status in this cohort. This finding is encouraging from a potential clinical implementation perspective as it suggests that useful predictive information may be obtained directly from routinely acquired T1w MRI without requiring additional steps and workload to have detailed tumour delineation.

Previous radiomics studies have reported higher performance for HER2 prediction, including an ROC-AUC of approximately 0.87 in a test set using multiparametric MRI and lesion-based radiomic signatures \citep{luo2022radiomics} and 0.74 using machine-learning analysis of multiparametric MRI features \citep{heitkamp2023receptor}. However, these approaches used multiple MRI sequences and substantially richer lesion-level feature sets than the present tumour-burden baseline. In the first case, Luo \textit{et al.} also studied a relatively small, single-centre cohort of patients undergoing brain metastasis resection, while, in the second case, Heitkamp \textit{et al.} performed lesion-level analysis using up to 10 metastases per patient and a large set of handcrafted features \citep{luo2022radiomics,heitkamp2023receptor}. Although the presented SFCN performance was lower than those reported by these previous multiparametric radiomics approaches, the simpler input requirements may offer advantages for workflow integration and generalizability, particularly where expert lesion segmentation is unavailable.

Several limitations of this work should be considered. First, the cohort was relatively small, comprising 135 patients from a single publicly available dataset. Second, the study was retrospective and lacks independent external validation. Third, the temporal relationship between the MRI examination and the tissue-derived HER2 label may introduce biological mismatch if receptor status changed between the two time points. Finally, the present study evaluates only post-contrast T1w MRI scans and did not incorporate other potentially informative imaging sequences or clinical variables.

Future work should focus on improving segmentation-free prediction from routinely acquired MRI scans and validating the approach in larger, independent, multi-center cohorts, particularly to include datasets in which repeat biomarker is available from biopsied or resected brain metastases to serve as the ground truth. Additionally, with the advent of HER-2 low status as a separate important biological entity aside from just HER-2 positive vs. negative, further discriminative evaluation of HER-2 status in breast cancer brain metastases is warranted. Automated tumour segmentation models (\textit{e.g.}, MedSAM \citep{ma2024segment}) could also be investigated as an intermediate strategy if explicit segmentation consistently improves performance while reducing the need for manual contouring. Additional MRI sequences, clinical variables, treatment history, and prior HER2 status may provide complementary information. A particularly important future comparison would be whether MRI-based prediction provides information beyond the HER2 status of the primary tumour or other available metastatic tissue.

Overall, these findings support the use of deep learning as a potential alternative to predefined lesion-level radiomics for investigating HER2 status in BCBM. The modest, non-significant difference between segmentation-free and mask-guided SFCN performance is particularly encouraging for future development of lower-workflow approaches that operate directly on whole-brain T1w MRI scans.

\section{Conclusion}

Predicting HER2 status from structural brain MRI scans alone is a challenging task, but the present results demonstrate a measurable imaging-associated signal that can be learned using deep learning methods. The relatively small, non-significant improvement associated with providing tumour segmentation masks suggests that much of the predictive signal captured by the SFCN model can potentially be learned from whole-brain T1w MRI scans alone. These findings support further investigation of segmentation-free deep-learning approaches for non-invasive, complementary HER2 assessment in BCBM, while recognizing that substantially larger cohorts and independent validation are required before clinical application.

\bibliographystyle{vancouver}
\bibliography{references}

\end{document}